\pdfoutput=1

\documentclass[11pt]{article}

\usepackage{acl}

\usepackage{times}
\usepackage{latexsym}

\usepackage[T1]{fontenc}

\usepackage[utf8]{inputenc}

\usepackage{microtype}

\usepackage{graphicx}
\usepackage{caption}
\usepackage{subcaption}
\usepackage{arydshln}
\title{Cited Text Spans for Scientific Citation Text Generation}

\author{Xiangci Li ~~ Yi-Hui Lee ~~ Jessica Ouyang\\
  Department of Computer Science \\
  University of Texas at Dallas \\
  Richardson, TX 75080 \\
  \tt lixiangci8@gmail.com, \\ 
  \tt \{yi-hui.lee, Jessica.Ouyang\}@UTDallas.edu
}

\begin{document}
\maketitle
\begin{abstract}
An automatic citation generation system aims to concisely and accurately describe the relationship between two scientific articles. To do so, such a system must ground its outputs to the content of the cited paper to avoid non-factual hallucinations. Due to the length of scientific documents, existing abstractive approaches have conditioned only on cited paper \textit{abstracts}. We demonstrate empirically that the abstract is not always the most appropriate input for citation generation and that models trained in this way learn to hallucinate. We propose to condition instead on the \textit{cited text span} (CTS) as an alternative to the abstract. Because manual CTS annotation is extremely time- and labor-intensive, we experiment with distant labeling of candidate CTS sentences, achieving sufficiently strong performance to substitute for expensive human annotations in model training, and we propose a human-in-the-loop, keyword-based CTS retrieval approach that makes generating citation texts grounded in the full text of cited papers both promising and practical.
\end{abstract}

\begin{figure*}[t]
\centering
  \includegraphics[width=\textwidth]{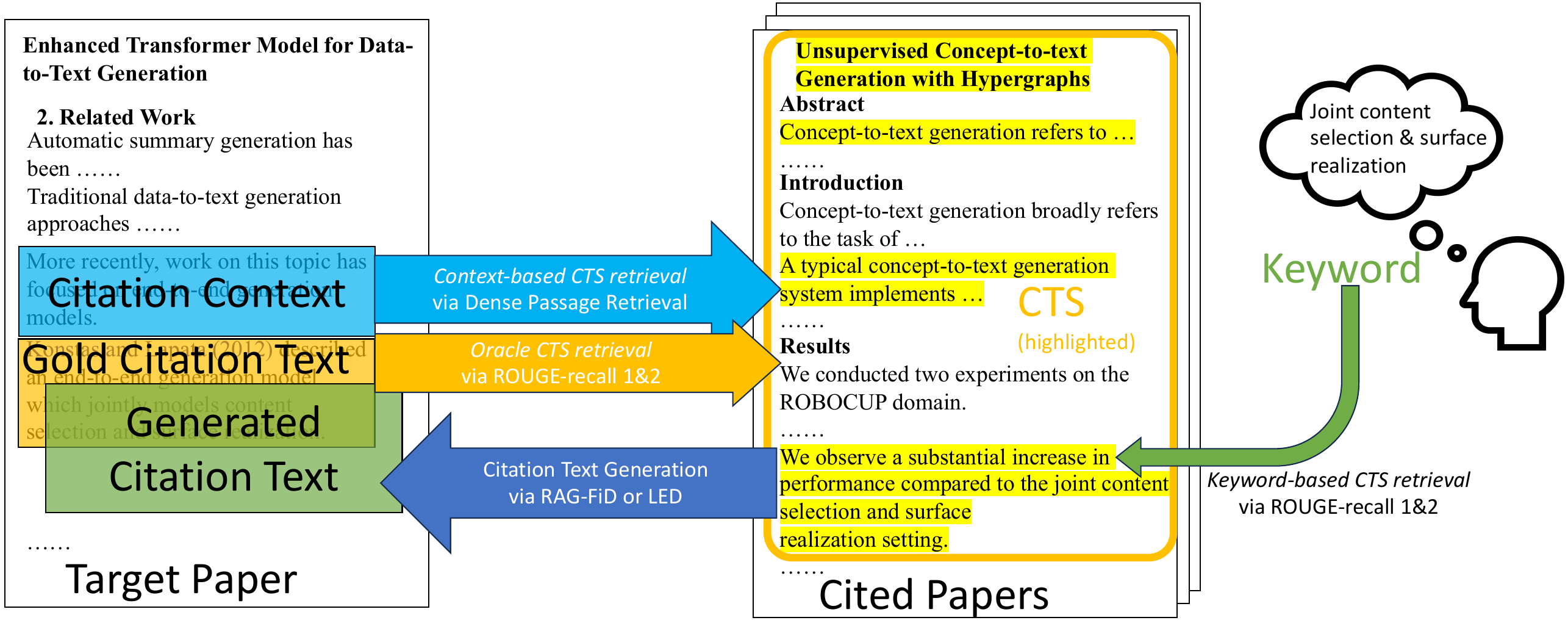}
  \vspace{-2em}
  \caption{Overview of the proposed CTS-based citation generation approach. The \emph{Context}, \emph{Oracle}, and \emph{Keyword} strategies are used to retrieve CTS from the cited paper (\citet{konstas-lapata-2012-unsupervised}) and generate a citation text for the target paper (\citet{gong-etal-2019-enhanced}). See Figure \ref{fig:example} for details of the example.} 
  \label{fig:illustration}
  \vspace{-0.5em}
\end{figure*}

\section{Introduction}
An automatic citation generation system aims to concisely and accurately describe the relationship between two scientific articles, a citing paper and a cited paper. Due to the rigorous nature of academic research, even large language models (LLMs), such as the GPT family \cite{brown2020language,openai2023gpt4}, still have to ground their outputs to the input cited papers to avoid non-factual hallucinations.

Given a set of papers to cite, prior extractive approaches \cite{hoang-kan-2010-towards, hu-wan-2014-automatic, chen2019automatic, wang2019toc, deng2021automatic} selected salient sentences from the full text of the cited papers. However, existing abstractive approaches \cite{abura2020automatic, xing-etal-2020-automatic, ge-etal-2021-baco, luu-etal-2021-explaining, chen-etal-2021-capturing, li-etal-2022-corwa} generate citation texts conditioned on only cited paper \textit{abstracts}, since the full texts of the papers often exceed the length limit of fine-tuned Transformer-based models \cite{li2022automatic}. \citet{li-etal-2022-corwa} show that, when human judges are asked to rate the \emph{relevance} of human-written gold citations with respect to the cited paper abstract (i.e. the generation system input), the scores are significantly lower than those of system-generated citations. We hypothesize that this discrepancy is caused by the gold citations referring to content from the cited paper body sections, not just the abstract, and these body sections are not available to either the generation model or the human judges.

Intuitively, the \textit{cited text span} (CTS), the specific span of the cited paper that the gold citation refers to, should be a better alternative to the abstract for grounding a citation. Unfortunately, CTS labels are difficult to obtain. Existing manually-annotated CTS datasets \cite{jaidka2018insights, jaidka2019cl, aburaed-etal-2020-multi} suffer from two key issues: annotation is extremely labor-intensive, even for domain experts, so the datasets are small; and inter-annotator agreement is relatively low (Cohen's $\kappa$ of 0.16$\sim$0.52). Further, automatic CTS retrieval systems trained on these datasets give low performance (F1$<$0.2) and also use the gold citation text as input, making them inapplicable in a citation generation setting. Given these obstacles, CTS-based citation text generation is under-explored.

In this work, we study scalable CTS retrieval for citation text generation without expensive manual annotations. Our main contributions are as follows:
\begin{itemize}
    \item We argue that CTS retrieval is an important and under-explored subtask of citation text generation. We demonstrate that citation texts generated using CTS are significantly more accurate and faithful than those using the existing approach of cited paper abstracts.
    \item Because existing human-annotated CTS datasets are too small to train an effective CTS retrieval system, we show that distant CTS labeling achieves sufficiently good performance to substitute for expensive human annotations. Using this method, we create a large-scale, distantly-labeled CTS dataset. 
    \item We argue that existing CTS retrieval systems are inappropriate for the citation generation task because they use the gold citation text as the retrieval query. Instead, we propose a simple, human-in-the-loop, keyword-based CTS retrieval method that makes generating citations grounded in the full text of the cited paper both promising and practical\footnote{\url{https://github.com/jacklxc/CTS4CitationTextGeneration}}. 
\end{itemize}

\section{Background and Related Work}
\label{sec:previous_cts}
There are two main CTS datasets: \citet{jaidka2018insights, jaidka2019cl} proposed the CL-Scisumm shared task and manually annotated CTS, citation facets, and summaries for 364 citation texts that cite a set of 20 papers; \citet{aburaed-etal-2020-multi} manually annotated sentence-level CTS for 20 related work sections. Unfortunately, both suffer from low inter-annotator agreement, and automatic CTS retrieval systems trained and evaluated on these datasets perform poorly \cite{jaidka2018insights, jaidka2019cl,aburaed-etal-2020-multi,chandrasekaran2020overview}. 

Only a few studies have used CTS for scientific document summarization. ScisummNet \cite{yasunaga2019scisummnet} 
leveraged a ROUGE-based CTS retrieval approach \cite{cao2015ranking} to collect community-based evidence for summarizing a paper. \citet{wang2019toc} trained a feature-based ensemble model for CTS retrieval using ScisummNet and used the retrieved CTS for extractive related work generation. \citet{syed-etal-2023-citance} used BM25 and SciBERT \cite{beltagy-etal-2019-scibert} to retrieve CTS to provide more contexts for scientific paper summarization. However, all of these works used the target citation text as the query for CTS retrieval, which cannot be used in the citation generation setting that we tackle in this paper. 


\section{A Large-Scale CTS Dataset: Distant vs. Human Labeling} \label{sec:cts_retrieval}

When authors cite a paper, they locate salient passages and compose a citation text grounded in these cited text spans (CTS). 
Unfortunately, such fine-grained CTS information is not recorded anywhere; even an author may not be able to precisely identify the CTS that they used. While prior CTS retrieval systems have trained on small, human-annotated CTS datasets \cite{jaidka2018insights, jaidka2019cl,aburaed-etal-2020-multi,chandrasekaran2020overview}, we argue that the relatively low inter-annotator agreement of these datasets suggests there can be multiple reasonable CTS for a given citation, just as there can be multiple valid extractive summaries of a document (see Section \ref{sec:manual_CTS_findings}). Thus, distantly-labeled CTS is not necessarily of lower quality than human-annotated CTS, as long as the distant labels identify one of these multiple reasonable candidates. In this section, we use a simple distant approach, with the gold citation text as query, to automatically label plausible CTS candidates to create a large-scale CTS dataset without expensive and time-consuming human annotations. We demonstrate that our distantly-labeled CTS are of comparable quality to existing human-annotated datasets. 


\subsection{Approach} \label{sec:rouge_retrieval}

\begin{figure}[t]
\centering
  \includegraphics[width=0.45\textwidth]{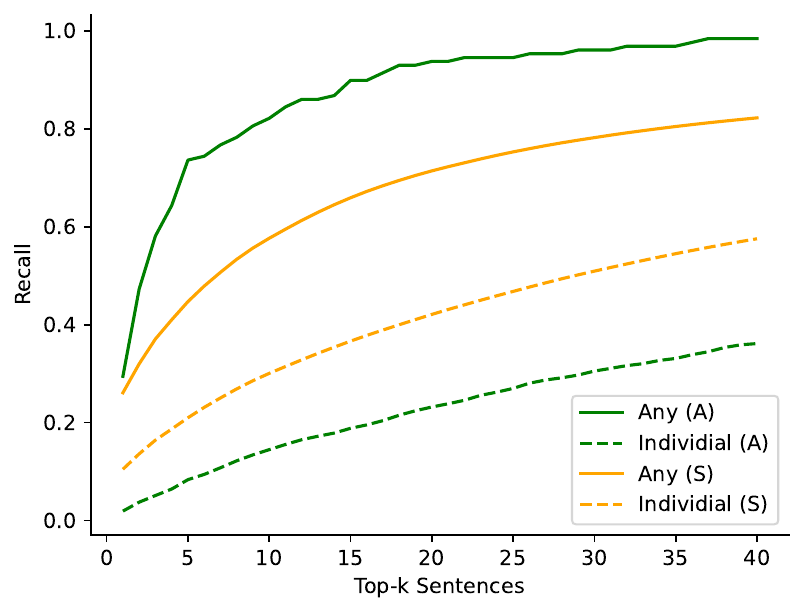}
  \vspace{-0.5em}
  \caption{Performance of distantly-labeled CTS, measured by recall against human-labeled CTS. Solid lines (``Any''): each citation counts as one data point; a true positive is when at least one human-labeled sentence is distantly-labeled. Dotted lines (``Individual''): each human-labeled sentence is a separate data point. Green lines (``A"): AbuRa'ed. Yellow lines (``S"): CL-SciSumm.} 
  \label{fig:retrieval_vs_annotaiton}
  \vspace{-1em}
\end{figure}

With the observation that high keyword overlap with the citation text is a crucial characteristic of CTS, we follow \citet{cao2015ranking, yasunaga2019scisummnet} in using gold citation texts as queries to rank the top-$k$ sentences in the corresponding cited papers as CTS candidates. The similarity is measured by the average of the ROUGE-1 and -2 recall scores \cite{lin-2004-rouge}.\footnote{We also experiment with BERTScore \cite{bert-score} as a dense retrieval approach, but the latter slightly underperforms ROUGE in all our experiments, including the comparison to human-labeled CTS (Figure \ref{fig:retrieval_vs_annotaiton}).} We use NLTK\footnote{\url{https://www.nltk.org/index.html}} to remove stop words for both the query and cited paper sentences. Compared to dense retrievers \cite{bert-score}, ROUGE-based retrieval does not use GPUs, and thus computationally more efficient. Moreover, since there is no good fine-tuning dataset for CTS retrieval due to the small size of human-labeled datasets \citet{jaidka2018insights, jaidka2019cl}, an unsupervised approach is preferable. 

We use the human-labeled datasets of \citet{jaidka2018insights, jaidka2019cl} (CL-SciSumm) and \citet{aburaed-etal-2020-multi} as test sets to evaluate our distantly-labeled CTS\footnote{We exclude 45 of \citet{aburaed-etal-2020-multi}'s 239 cited papers due to an XML format error, but we do not expect this omission to affect the overall trend of our evaluation.}. 
We take the union of all annotators' selected sentences as the human-labeled CTS. 

\subsection{Evaluation Metrics}

\paragraph{Token overlap.} We use ROUGE-recall to measure the coverage of the distantly-labeled tokens that support the gold citation text.

\paragraph{Faithfulness.} We use QuestEval \cite{scialom-etal-2021-questeval} and ANLI \cite{nie-etal-2020-adversarial} to compare the faithfulness of the gold citation text when grounded in human-labeled versus distantly-labeled CTS. For ANLI, we report the non-contradiction score (sum of \emph{entailment} and \emph{neutral}) given by a fine-tuned RoBERTa-large model \cite{liu2019roberta}.

\label{sec:automatic_vs_manual_results}

\begin{figure}[t]
\centering
  \includegraphics[width=0.45\textwidth]{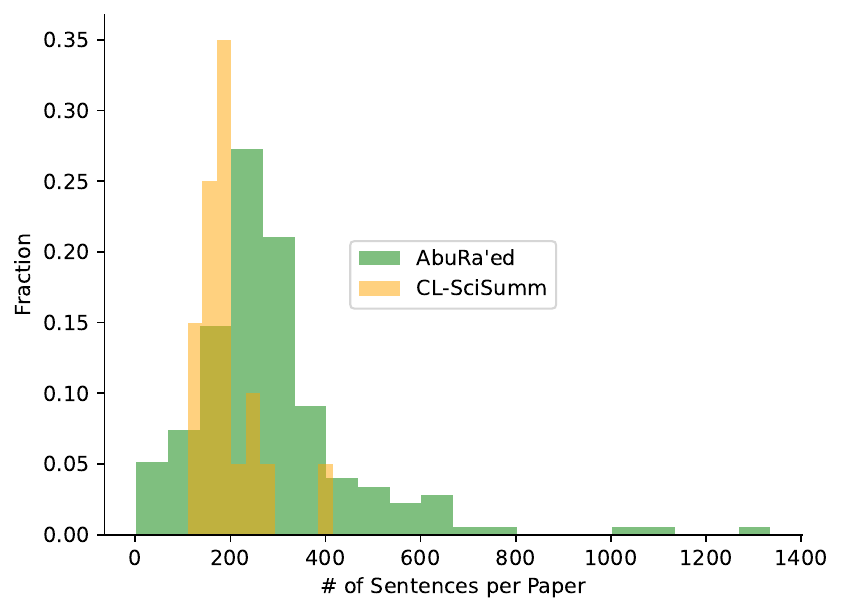}
  \vspace{-0.5em}
  \caption{Distribution of the lengths of the cited papers by the number of sentences in CL-SciSumm and AbuRa'ed.} 
  \label{fig:paper_lengths}
  \vspace{-1em}
\end{figure}

\subsection{Findings} \label{sec:manual_CTS_findings}
\paragraph{Distant labeling has high coverage of human annotations.} 
As Figure \ref{fig:retrieval_vs_annotaiton} shows, our top-40 retrieved CTS cover ~80\% and ~95\% of the human-labeled CTS in CL-SciSumm and AbuRa'ed, respectively. Further, as Figure \ref{fig:paper_lengths} shows, the papers in these two datasets average 200 sentences long, so we can see that our automatic ranking concentrates the human-annotated CTS into roughly the top 20\% ($\sim$40/200) of cited paper sentences.

\begin{figure}
\centering
  \includegraphics[width=0.45\textwidth]{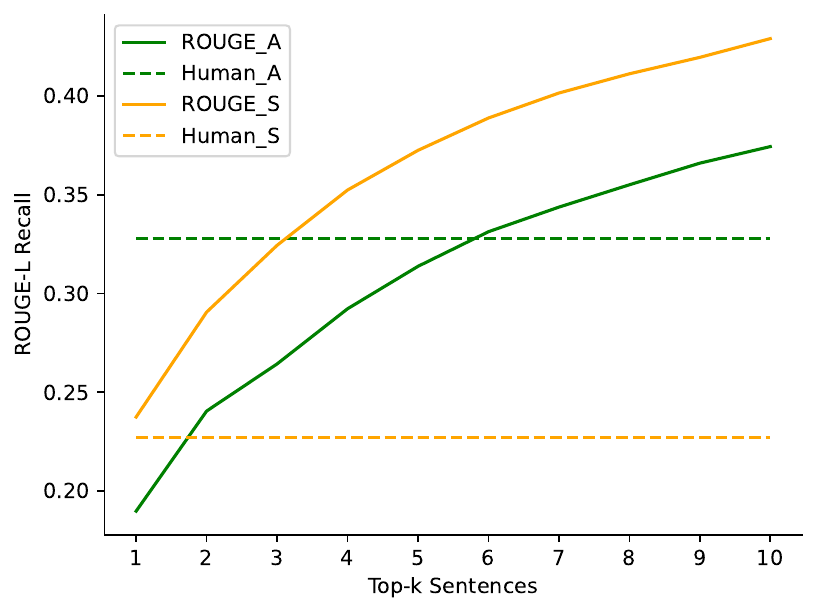}
  \vspace{-0.5em}
  \caption{Overlap of the top-$k$ distantly-labeled CTS (solid lines) \& human-labeled CTS (dotted lines), measured by ROUGE-L recall against the gold citation text. Green lines (``A"): AbuRa'ed. Yellow lines (``S"): CL-SciSumm. The average length of human-labeled CTS is $|\overline{A}|$=16.8 and $|\overline{S}|$=2.5 sentences.} 
  \label{fig:human_annotation_vs_target}
  \vspace{-0.5em}
\end{figure}

\begin{table}[t]
\begin{center}
\setlength{\tabcolsep}{3pt} 
\renewcommand{\arraystretch}{1.0} 
\small
    \begin{tabular}{  l | l l l | l l l }
    \hline
     & \multicolumn{3}{l}{\textit{Distantly-Labeled}} & \multicolumn{3}{l}{\textit{Human-Labeled}} \\
    \textbf{Config} & \textbf{R-L} & \textbf{QEval} & \textbf{ANLI} & \textbf{R-L} & \textbf{QEval} & \textbf{ANLI}\\ \hline
    AbuRa'ed & \textbf{0.376} & 0.254 & 0.959 & 0.328 & 0.263 & \textbf{1.000} \\ 
    A-strict  & - & - & - & 0.224 & \textbf{0.269} & 0.992 \\ 
    \hline
    SciSumm  & \textbf{0.429} & 0.259 & \textbf{0.973} & 0.225 & 0.258 & 0.922 \\ 
    S-strict  & - & - & - & 0.187 & \textbf{0.275} & 0.923 \\ 
    \hline
    \end{tabular}
    \vspace{-0.5em}
    \caption{Distant and human CTS labeling performance measured by ROUGE-L-recall, QuestEval, and ANLI against the gold citation text. ``Strict" uses the intersection, rather than union, of human-labeled sentences by different annotators. $k=10$ for distant-labeling.} 
    \label{tab:human_annotation_vs_target}
    \vspace{-1.5em}
\end{center}
\end{table}

\paragraph{Distant labeling is more similar to the gold citation text.}
\label{sec:rouge_vs_human_annotation}

As Figure \ref{fig:human_annotation_vs_target} and Table \ref{tab:human_annotation_vs_target} show, although both human-labeled and distantly-labeled CTS are comparably faithful, distantly-labeled CTS has higher token overlap with the target citation text. This is unsurprising, given that using ROUGE as our similarity metric explicitly prioritizes token overlap, while human annotators may not do so. Evaluating human CTS in the intersection, rather than the union, of different annotators' labels (\emph{strict}) has a noticeable impact on the token overlap, indicating that \emph{individual annotators can select different, but all reasonable, CTS sentences}.

\paragraph{Distant labeling performs better on the downstream generation task.}

\begin{table}[t]
\begin{center}
\setlength{\tabcolsep}{1pt} 
\renewcommand{\arraystretch}{1.0} 
\small
    \begin{tabular}{  l | ccc | ccc }
    \hline
     & \multicolumn{3}{c}{\textit{Distantly-Labeled}} & \multicolumn{3}{c}{\textit{Human-Labeled}} \\
    \textbf{Config} & \textbf{BLEU} & \textbf{METEOR} & \textbf{R-L} & \textbf{BLEU} & \textbf{METEOR} & \textbf{R-L}\\ \hline
    AbuRa'ed  & 0.042 & 0.206 & 0.220 & 0.052 & 0.216 & 0.201 \\
    A-strict  & - & - & - & \textbf{0.058} & \textbf{0.228} & \textbf{0.226} \\ \hline
    SciSumm  & \textbf{0.064} & \textbf{0.291} & \textbf{0.265} & 0.061 & 0.252 & 0.242 \\ 
    S-strict  & - & - & - & 0.058 & 0.236 & 0.250 \\ 
    \hline
    \end{tabular}
    \vspace{-0.5em}
    \caption{Citation text generation performance measured by BLEU, METEOR and ROUGE-F1 against the gold citation. ``Strict": intersection, rather than union, of human-labeled sentences.} 
    \label{tab:sentence_generation_performance}
    \vspace{-1.5em}
\end{center}
\end{table}

We also experiment with citation text generation using both distantly-labeled and human-labeled CTS, using the \emph{LED-oracle} model described in Section \ref{sec:citation_text_generation}. As Table \ref{tab:sentence_generation_performance} shows, distantly-labeled CTS performs comparably to human-labeled CTS for the AbuRa'ed dataset and outperforms human labels for CL-SciSumm, confirming that distant CTS labeling produces competitive performance on the downstream generation task. This experiment shows the upper bound of CTS performance using human and distant oracles, demonstrating that \textit{it is not necessary to acquire expensive, human-labeled CTS for training a retriever for downstream citation text generation}.  


\section{Citation Text Generation Using CTS} \label{sec:citation_text_generation}

In this section, we experiment with citation text generation conditioned on CTS, rather than the abstract-only approach used in prior work, and demonstrate that conditioning on CTS produces more accurate and faithful citations.

\subsection{Problem Formulation}
Following \citet{li-etal-2022-corwa}, who note the information loss and leak issue of the citation \textit{sentence} generation task, we aim to generate citation \textit{spans} conditioned on (1) the citing paper context and (2) the full text of the cited paper(s) via retrieved CTS. We assume an auto-regressive generation setting, where citations are generated one at a time, in order of their appearance in the citing paper. Thus, we have access to up to two sentences before the target citation to use as context.

\paragraph{Data.}

We use the CORWA citation span\footnote{To avoid confusion between cited text \textit{spans} and citation \textit{spans}, we will refer to the generation targets as \textit{citation texts}.} dataset \citet{li-etal-2022-corwa}, which is derived from the ACL partition of S2ORC \cite{lo-etal-2020-s2orc}. CORWA consists of a human-annotated \textit{training} set and \textit{test} set, as well as an automatically-labeled \textit{distant} training set, with 1654, 1206, and 19784 ``dominant"-type citation texts, respectively. We use only the dominant-type citations that are grounded to specific CTS, following \citet{jaidka2018insights, jaidka2019cl,aburaed-etal-2020-multi}, who filter out very short (``reference"-type) citations.

\subsection{Approach} \label{sec:approach}

\subsubsection{CTS Retrieval Strategies}  \label{sec:cts_retreival}
\paragraph{Context-based retrieval.} 
In this \textit{fully automatic} approach, we retrieve CTS given only the target citation's context sentences. 
We apply the distant labeling approach described in Section \ref{sec:cts_retrieval} on the CORWA training set to create a large-scale CTS dataset and use this data to train a Dense Passage Retrieval \citep[DPR;][]{karpukhin-etal-2020-dense} model. We initialize both the query and candidate document encoders with Aspire \cite{mysore-etal-2022-multi}, a scientific sentence encoder fine-tuned on top of SciBERT \cite{beltagy-etal-2019-scibert}. For training, we use the top-40 distantly-labeled CTS sentences as positive documents and the rest of the sentences in the cited papers as hard-negative documents. 

\paragraph{Unsupervised oracle retrieval.}
The distant labeling method described in Section \ref{sec:cts_retrieval} uses the gold citation text as a query; while this approach cannot be used at test time when the target citation text is not known, we use it as an \emph{Oracle} setting to explore the upper-bound of CTS-based citation text generation. 

\paragraph{Human-in-the-loop keyword retrieval.} \label{sec:human-in-the-loop}
There may be multiple reasonable citation texts given the same input context, depending on the interests of the reader. Therefore, we also experiment with a practical, human-in-the-loop scenario where a researcher already has a general idea of what they want to know about a particular cited paper, so they input some keywords as a query to guide CTS retrieval. To simulate this scenario, we apply a keyword extractor to the target citation texts and use the extracted keywords as simulated user-input keywords to retrieve CTS using unsupervised retrieval (\emph{Keyword}). See Figure \ref{fig:example} for an example.

To train the keyword extractor, we manually annotate a small set (673 instances) of citation keywords on top of the CORWA dataset. Each instance consists of a short, ``reference"-type citation text with manually highlighted keywords\footnote{We use reference-type citations to train the keyword extractor only, not in any citation text generation experiments.}. We pre-train a T5-small model \cite{raffel2020exploring} on the named-entity recognition labels of the SciREX training set \cite{jain-etal-2020-scirex} and further fine-tune it on manually annotated CORWA keywords as a sequence-to-sequence keyword extractor\footnote{Alternatively, any reasonable keyword extractor, such as recent LLMs, can also be used to simulate the user-input keywords.}. 

\paragraph{} Table \ref{tab:CTS_retrieval_performance} shows the token overlap (ROUGE-L recall) between CTS retrieved using each of our three strategies and the target citation, as well as the faithfulness of the target citation grounded in the CTS; we compare against the cited paper abstract, which is used by existing citation generation systems instead of CTS. The token overlap of CTS is much higher than that of the abstract, as we expect, and more importantly, the faithfulness of the target citation grounded in CTS is higher than if it were grounded in the abstract, supporting our hypothesis that the human-written target citations use information from the body sections of the cited papers. Thus, it is important to condition on CTS, not just the cited paper abstract, for citation generation.


\begin{table}[t]
\begin{center}
\setlength{\tabcolsep}{5pt} 
\small
    \begin{tabular}{  l | l l l }
    \hline
    \textbf{Setting} & \textbf{ROUGE-L} & \textbf{QuestEval} & \textbf{ANLI} \\ \hline
    Abstract & 0.290  & 0.263 & 0.868 \\ \hline
    Context CTS & 0.400 & 0.265 & \textbf{0.964} \\ 
    Oracle CTS  & \textbf{0.533} & \textbf{0.271} & 0.955\\ 
    Keyword CTS & 0.440 & \textbf{0.271} & 0.949 \\
    \hline
    \end{tabular}
    \vspace{-0.5em}
    \caption{Quality of citation generation input in terms of token coverage (ROUGE-recall) and faithfulness (QuestEval, ANLI) with respect to the target citation text.} \label{tab:CTS_retrieval_performance}
    \vspace{-1.5em}
\end{center}
\end{table}


\subsubsection{Citation Text Generation Strategies}

\paragraph{Retrieval-Augmented Generation.}
The Retrieval-Augmented Generation \citep[RAG;][]{lewis2020retrieval} bridges document retrieval and text generation by fine-tuning the query encoder and generator jointly. RAG uses DPR \cite{{karpukhin-etal-2020-dense}} to retrieve $k$ documents using the input query. We concatenate the target citation's context sentences and the retrieved CTS sentences\footnote{Unless explicitly specified, to balance performance and GPU memory usage, we retrieve the top $k=10$ CTS in all experiments for the inference of citation text generation.} as the input to the citation text generator. We use Fusion-in-Decoder \citep[FiD;][]{izacard-grave-2021-leveraging} with BART-base \cite{lewis-etal-2020-bart} as the generator in our experiments. 

\paragraph{Longformer-Encoder-Decoder.}
We also experiment with a simple Longformer-Encoder-Decoder \citep[LED;][]{beltagy2020longformer} model. We input the concatenation of the target citation context, the citation mark (eg. ``Smith et al. (2024)"), and the retrieved CTS for each cited paper.

\subsubsection{Experimental Settings}


\begin{table}[t]
\begin{center}
\setlength{\tabcolsep}{1pt} 
\renewcommand{\arraystretch}{1.0} 
\small
    \begin{tabular}{  l | ccc | ccc }
    \hline
     & \multicolumn{3}{c}{\textit{RAG-FiD}} & \multicolumn{3}{c}{\textit{LED}} \\
    \textbf{Setting} & \textbf{BLEU} & \textbf{METEOR} & \textbf{R-L} & \textbf{BLEU} & \textbf{METEOR} & \textbf{R-L}\\ \hline
    Abstract & 0.214 & 0.397 & 0.194 & 0.142 & 0.380 & 0.189 \\ \hline
    Context & 0.216 & 0.397 & 0.197 & 0.163 & 0.359 & 0.194 \\
    Oracle & \textbf{0.258} & \textbf{0.461} & \textbf{0.292} & \textbf{0.216} & \textbf{0.450} & \textbf{0.312} \\ 
    Keyword & \textit{0.239} & \textit{0.437} & \textit{0.262} & \textit{0.182} & \textit{0.404} & \textit{0.270} \\ 
    \hline
    \end{tabular}
    \vspace{-0.5em}
    \caption{Citation text generation performance measured by BLEU, METEOR and ROUGE-F1 against the gold citation. Citation marks are excluded in evaluation.} 
    \label{tab:span_generation_performance}
    \vspace{-1.5em}
\end{center}
\end{table}

\begin{table}[t]
\begin{center}
\setlength{\tabcolsep}{1pt} 
\renewcommand{\arraystretch}{1.0} 
\small
    \begin{tabular}{  l | l l l l }
    \hline

    \textbf{Setting} & \textbf{Fluency} & \textbf{Relevance} & \textbf{Coherence} & \textbf{Overall}\\ \hline
    Gold & 4.71 (4.74) & 4.04 (4.18) & \textbf{4.23 (4.41)} & 3.87 \textbf{(4.16)} \\ \hline
    Abstract & 4.71 (4.72) & \textbf{4.29 (4.26)} & 4.06 (3.96) & 3.77 (3.63) \\ \hline
    Context & 4.80 (4.80) & 4.41 (4.31) & 4.00 (3.96) & \textbf{3.95} (3.85) \\
    Oracle & 4.80 (4.80) & 4.18 (4.14) & 4.07 (4.09) & 3.86 (3.80) \\
    Keyword & \textbf{4.84 (4.84)} & 4.21 (4.16) & 4.07 (4.07) & 3.84 (3.76) \\
    \hline
    \end{tabular}
    \vspace{-0.5em}
    \caption{Human evaluation with moderate inter-annotator agreement (avg. Kendall's $\tau = 0.239$). Scores in parentheses are from the second round after the gold citation texts are revealed to the judges. } \label{tab:human_evaluation}
    \vspace{-1.5em}
\end{center}
\end{table}

\paragraph{Training.}
We train on the large, automatically-labeled CORWA \textit{distant} training set using cross-entropy loss and perform hyper-parameter tuning on the smaller, human-labeled \textit{training} set; details and the hyper-parameter settings are reported in Appendix \ref{sec:more_details}. Each model takes about one day to train on one Nvidia Tesla V100S-32GB GPU.

\paragraph{Testing.} 
For all experiments, we report performance on the human-labeled CORWA \textit{test} set. 


\subsection{Human Evaluation}

Following the setup used in prior works \cite{xing-etal-2020-automatic, ge-etal-2021-baco, li-etal-2022-corwa}, we recruit six NLP graduate students who are fluent in English as human evaluation judges. To increase their coverage of the test set, we split the judges into two groups of three and assign each group 25 instances sampled from the output of the RAG-FiD model. Each instance consists of a single target citation context and several candidate generations: one conditioned on the \emph{Abstract} only and one for each of our three CTS retrieval strategies: \emph{Context}, \emph{Oracle}, and \emph{Keyword}. The target citation context, cited paper abstracts, and retrieved CTS from all three strategies are also provided to the judges. 
Each generated citation is rated on a five-point Likert scale with respect to \textit{Fluency}, \textit{Relevance} (to the input; either abstract or CTS), \textit{Coherence} (with respect to the context), and \textit{Overall Quality}. 

We perform two rounds of evaluation: first, the gold citation texts are mixed in with the candidate generations but are not identified to the judges; second, we reveal the gold citation texts and ask the judges to revise their ratings with this new information.

\begin{table}[t]
\begin{center}
\setlength{\tabcolsep}{3pt} 
\renewcommand{\arraystretch}{1.0} 
\small
    \begin{tabular}{  l | l l | l l }
    \hline
     & \multicolumn{2}{l}{\textit{RAG-FiD}} & \multicolumn{2}{l}{\textit{LED}} \\
    \textbf{Setting} & \textbf{R-L} & \textbf{ANLI} & \textbf{R-L} & \textbf{ANLI}\\ \hline
    Abstract & 0.556 & 0.745  & 0.591 & 0.744  \\ \hline
    Context & 0.641 & \textbf{0.959} & 0.626 & \textbf{0.954}  \\
    Oracle & \textbf{0.726} & 0.940 & \textbf{0.682} & 0.950 \\ 
    Keyword  & 0.630 & 0.941 & 0.604 & 0.946  \\ 
    \hline
    \end{tabular}
    \vspace{-0.5em}
    \caption{Similarity and faithfulness of the generated citation texts with respect to their input setting, measured by ROUGE-recall and ANLI.} \label{tab:abstractive_vs_extractive_faithfullness}
    \vspace{-1.5em}
\end{center}
\end{table}

\subsection{Results and Discussion} \label{sec:results}

\begin{figure*}
     \centering
     \begin{subfigure}[b]{0.29\textwidth}
         \centering
         \includegraphics[width=\textwidth]{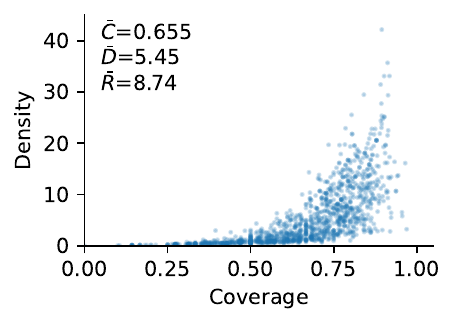}
         \vspace{-2.2em}
         \caption{RAG-FiD Abstract}
         \label{fig:fid_abstract}
     \end{subfigure}
     \quad
     \begin{subfigure}[b]{0.29\textwidth}
         \centering
         \includegraphics[width=\textwidth]{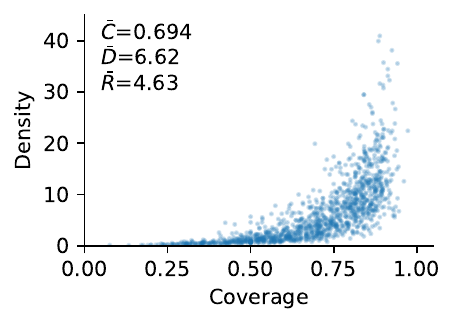}
         \vspace{-2.2em}
         \caption{LED Abstract}
         \label{fig:led_abstract}
     \end{subfigure}
     \quad
     \begin{subfigure}[b]{0.29\textwidth}
         \centering
         \includegraphics[width=\textwidth]{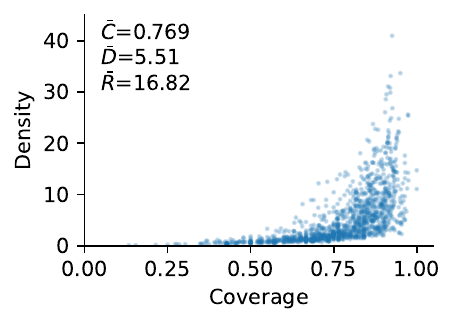}
         \vspace{-2.2em}
         \caption{RAG-FiD Context}
         \label{fig:fid_context}
     \end{subfigure}
     \quad
     \begin{subfigure}[b]{0.3\textwidth}
         \centering
         \includegraphics[width=\textwidth]{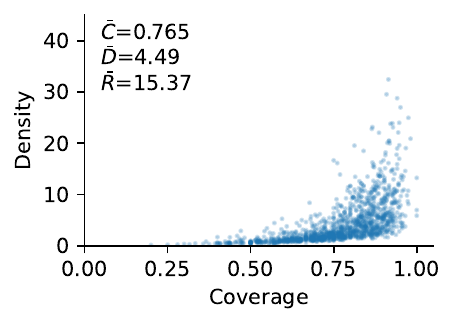}
         \vspace{-2.2em}
         \caption{LED Context}
         \label{fig:led_context}
     \end{subfigure}
     \quad
     \begin{subfigure}[b]{0.29\textwidth}
         \centering
         \includegraphics[width=\textwidth]{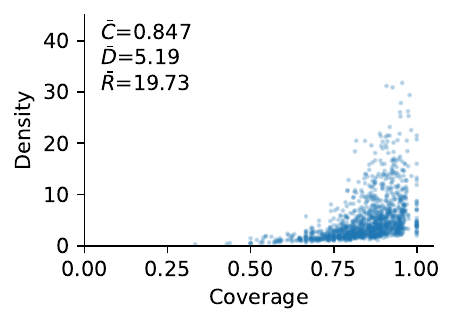}
         \vspace{-2.2em}
         \caption{RAG-FiD Oracle}
         \label{fig:fid_oracle}
     \end{subfigure}
     \quad
     \begin{subfigure}[b]{0.29\textwidth}
         \centering
         \includegraphics[width=\textwidth]{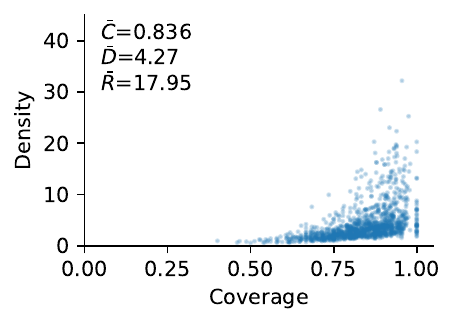}
         \vspace{-2.2em}
         \caption{LED Oracle}
         \label{fig:led_oracle}
     \end{subfigure}
     \quad
     \begin{subfigure}[b]{0.29\textwidth}
         \centering
         \includegraphics[width=\textwidth]{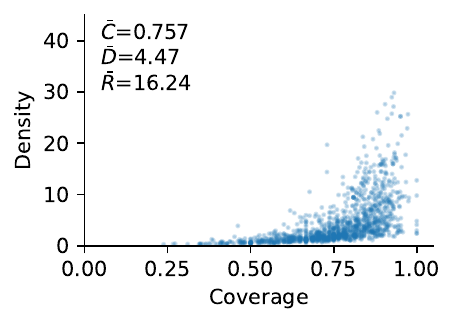}
         \vspace{-2.2em}
         \caption{RAG-FiD Keyword}
         \label{fig:fid_keyword}
     \end{subfigure}
     \quad 
     \begin{subfigure}[b]{0.29\textwidth}
         \centering
         \includegraphics[width=\textwidth]{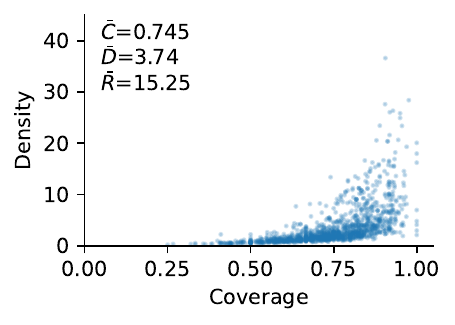}
         \vspace{-2.2em}
         \caption{LED Keyword}
         \label{fig:led_keyword}
     \end{subfigure}
     \quad
     \begin{subfigure}[b]{0.29\textwidth}
         \centering
         \includegraphics[width=\textwidth]{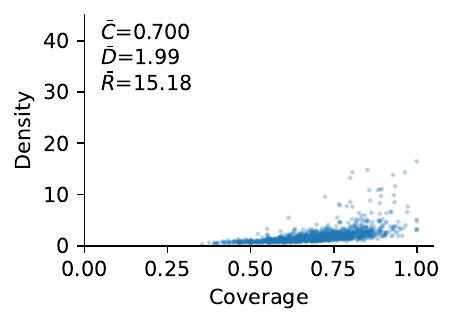}
         \vspace{-2.2em}
         \caption{Gold}
         \label{fig:led_target}
     \end{subfigure}     
     \vspace{-1.0em}
        \caption{Extractiveness of the generated citation texts, measured by \textit{coverage} and \textit{density} \cite{grusky-etal-2018-newsroom} against the generation input (abstract or CTS). The average coverage (C), density (D), and compression ratio (R) are shown for each generation setting. The ``Gold" setting (\ref{fig:led_target}) is measured using Oracle CTS as the ``input".}
        \label{fig:extractiveness}
\end{figure*}

\paragraph{CTS-based generation outperforms abstract-only generation.}
Table \ref{tab:span_generation_performance} shows that the \emph{Oracle} and \emph{Keyword} CTS settings massively outperform the \textit{Abstract} approach. Moreover, Table \ref{tab:human_evaluation} shows that \textit{Abstract} receives the lowest human \emph{Overall} rating. In contrast with prior works in the abstract-only setting \cite{abura2020automatic, xing-etal-2020-automatic, ge-etal-2021-baco, luu-etal-2021-explaining, chen-etal-2021-capturing, li-etal-2022-corwa}, all settings now score well on \emph{Relevance} (higher than 4). Previously, even the gold citation text scored poorly on \textit{Relevance} because the judges were only shown the abstract, which did not contain sufficient information to support the citation texts; our judges are shown the retrieved CTS sentences, which are a much better match for the citation texts. 

\paragraph{CTS-based generation extracts more short phrases.}
The ROUGE-recall scores in Table \ref{tab:abstractive_vs_extractive_faithfullness} show that, compared to \textit{Abstract}, CTS-based generations are more similar to their inputs, suggesting that models trained on CTS are more extractive. To investigate further, we use \citet{grusky-etal-2018-newsroom}'s \emph{coverage} and \emph{density} metrics. \emph{Coverage} measures how much of the generated citation text is extracted from the input (abstract or CTS), while \emph{density} measures the size of the extractions (e.g. individual unigrams versus entire sentences). Additionally, we calculate the compression ratio between the lengths of the inputs and the generated citation texts. 

As Figure \ref{fig:extractiveness} shows, although the \textit{Abstract} setting has the lowest compression ratio, it has a significantly lower \emph{coverage} score than the CTS settings, confirming that the latter is more extractive. From this finding, along with the lower faithfulness scores in Table \ref{tab:abstractive_vs_extractive_faithfullness}\footnote{We exclude QuestEval from this table due to an issue arising from the relative compression ratios of the different input settings. 
}, we can infer that \emph{Abstract} produces more non-factual hallucinations that distort the ideas of the cited paper. This finding supports our hypothesis that, in the abstract-only setting, \textit{the citation generation input does not contain all of the information present in the target gold citation text, forcing the model to learn to hallucinate that missing information}.

Moreover, \emph{Abstract} has a significantly higher \emph{density} score, indicating longer copied sequences from the input, while the CTS-based settings extract and reorganize shorter phrases from the input. Since copying long text spans from other documents results in plagiarism in the citation generation setting, CTS-based generation is again preferable. Nonetheless, we stress that our CTS-based generation approach is intended to increase accuracy and faithfulness and does not explicitly address the plagiarism issue; additional post-processing, such as a rewriting module, should be used to further reduce the risk of plagiarism.


\begin{table}[t]
\begin{center}
\setlength{\tabcolsep}{3pt} 
\renewcommand{\arraystretch}{1.0} 
\small
    \begin{tabular}{  l | l l l }
    \hline
    \textbf{Section} & \textbf{Oracle} & \textbf{Keyword} & \textbf{Context} \\ \hline
    introduction & 2155 (1) & 1681 (1) & 4140 (1) \\ 
    related work & 1217 (2) & 634 (5) & 401 (4) \\ 
    model or architecture  & 903 (3) & 1036 (2) & 383 (5) \\ 
    abstract  & 712 (4) & 816 (3) & 3667 (2) \\ 
    experimental results  & 710 (5) & 703 (4) & 92 (8) \\ 
    conclusion  & 508 (6) & 589 (6) & 316 (6) \\ 
    approach  & 312 (7) & 312 (8) & 141 (7) \\ 
    data  & 307 (8) & 370 (7) & 75 (9) \\ 
    evaluation  & 210 (9) & 215 (10) & 38 (12) \\ 
    discussion  & 88 (10) & 75 (11) & 15 (16) \\ 
    title  & 73 (11) & 262 (9) & 921 (3) \\ 
    background  & 67 (12) & 74 (12) & 54 (10) \\ 
    \hline
    \end{tabular}
    \vspace{-0.5em}
    \caption{Counts (and ranks) of the most common section headers of retrieved CTS for the CORWA test set. Approximately 12,000 CTS are retrieved in total.} \label{tab:retrieved_sections}
    \vspace{-1.5em}
\end{center}
\end{table}

\begin{figure*}
\centering
  \includegraphics[width=\textwidth]{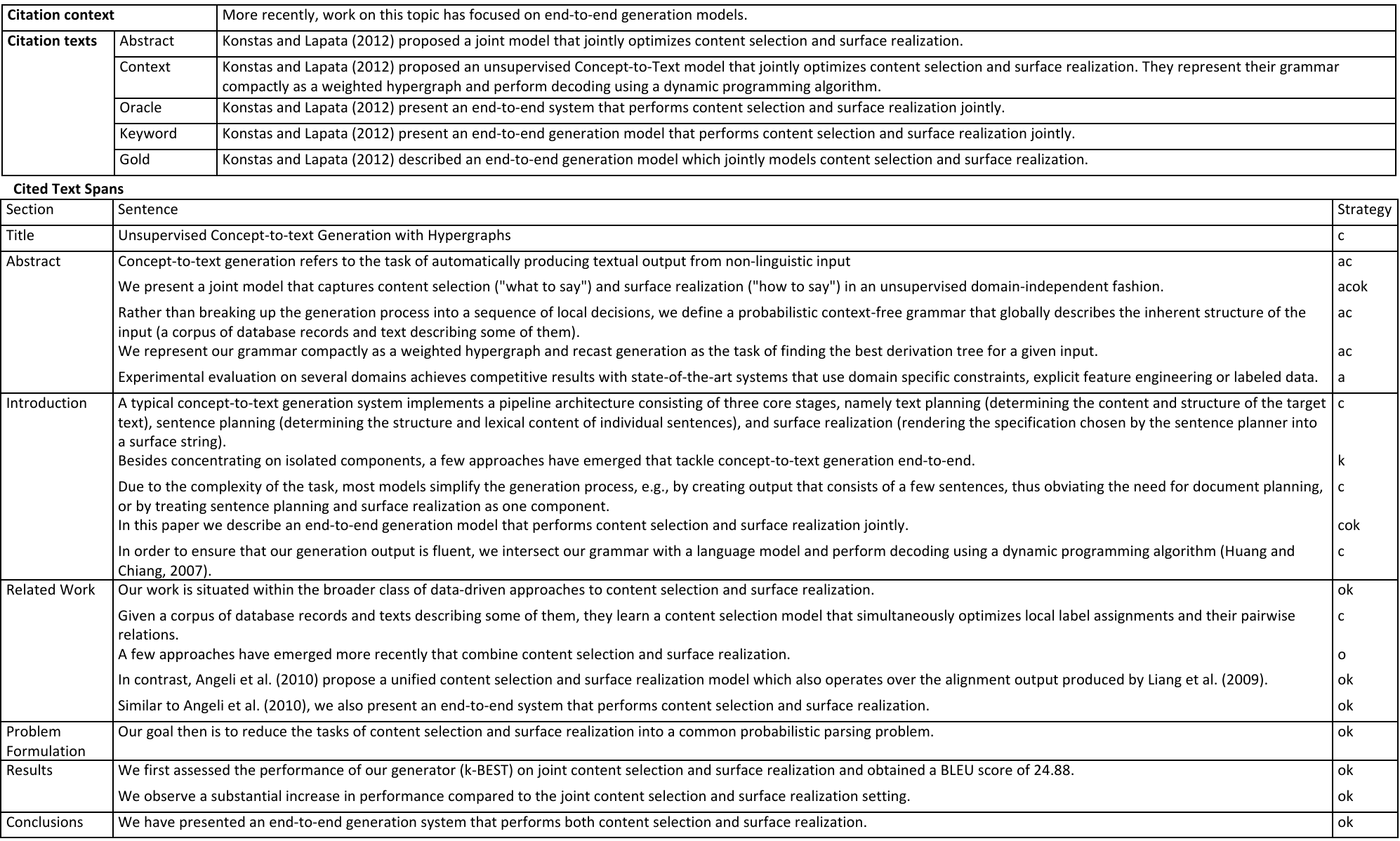}
  \vspace{-2em}
  \caption{Citation text generation example: \citet{gong-etal-2019-enhanced} citing \citet{konstas-lapata-2012-unsupervised}. The citation context is shown at the top, followed by the generated citation texts using the abstract and the three CTS strategies as inputs; the gold citation text is also shown. For each CTS sentence, ``strategy'' refers to the input setting(s) that included it: \textbf{a}bstract, \textbf{c}ontext CTS, \textbf{o}racle CTS, \textbf{k}eyword CTS. The keywords used are ``end-to-end generation model'', ``content selection'', and ``surface realization''.} 
  \label{fig:example}
  \vspace{-0.5em}
\end{figure*}

\paragraph{Predicting CTS fully automatically is challenging.}
Although the fully automatic \emph{Context} setting has strong input retrieval performance (Table \ref{tab:CTS_retrieval_performance}), its performance on the downstream citation text generation task is mixed. It underperforms on automated metrics (Table \ref{tab:span_generation_performance}), but human judges rate it slightly higher than \emph{Oracle} and \emph{Keyword} (Table \ref{tab:human_evaluation}). Table \ref{tab:retrieved_sections} shows that \emph{Oracle} and \emph{Keyword} tend to retrieve CTS from more specific sections, such as ``Model or Architecture'' or ``Results'' (Figure \ref{fig:example}), suggesting that the human judges may have felt that the generated citation texts contained too much unnecessary detail. 

Fully automatic CTS retrieval given only the target citation context sentences is challenging because the citation context and the CTS candidates come from different papers (citing versus cited), and the semantic mismatch between them lowers the performance of the DPR module. In the example in Figure \ref{fig:example}, the citation context only mentions ``end-to-end generation models'', omitting the ``content selection'' and ``surface realization'' keywords that appear in the gold citation text, making CTS retrieval in \textit{context} setting much harder than in \textit{oracle} and \textit{keyword}, which have explicit guidance with all important keywords from the gold citation text. 

Interestingly, Table \ref{tab:human_evaluation} shows that human evaluators cannot distinguish between gold and generated citation texts during the blind evaluation round. When the gold citation text is identified to the judges, they revise its scores higher, while lowering the scores of the generated candidates. This finding suggests that even humans need some guiding information to help them process citation texts. Thus, as \citet{li-etal-2022-corwa} argue, the citation text generation task should be human-in-the-loop, leveraging a researcher's expertise to reduce both the search space size and cascading error propagation in the pipeline, and our \textit{Keyword}-based CTS retrieval approach provides a practical workaround for the challenge of predicting CTS fully automatically.

\section{Conclusion}
In this work, 
we show that CTS-based citation text generation is more accurate and faithful than the abstract-based approaches used in prior work. We address the challenges that previously prevented CTS from being used effectively for citation text generation. Distant labeling, such as ROUGE-based labeling, achieves sufficiently strong performance to substitute for expensive manual CTS annotations, allowing us to build a dataset large enough to train a supervised CTS retriever. To avoid the pitfall of using the gold citation text as the retrieval query, which has been used in prior work but would not be possible in a real application, we propose a human-in-the-loop, keyword-based CTS retrieval approach that achieves strong performance in a practical and realistic setting. Finally, our experimental results show that, while retrieving CTS using only the citation context sentences is challenging and does not achieve the best performance using automated metrics, it is preferred by human judges and is a good direction for further investigation.

\section*{Limitations} \label{sec:discussion_conclusion}
\paragraph{Challenge of CTS retrieval.} As discussed in Section \ref{sec:previous_cts}, existing CTS retrieval approaches all use the gold citation text as the query. Our results also show that, while CTS retrieved using the gold citation text (or its keywords) yields strong generation performance, CTS retrieved with only the citation context sentences is not satisfactory. Therefore, we argue that citation text generation should be divided into two parts: support document retrieval and document-based citation text generation. Based on our results, we can achieve strong generation performance, but the automated retrieval part requires further investigation.

As a workaround solution without a strong, context-based CTS retriever, we show that, when the user provides a few keywords to guide CTS retrieval, our models can perform quite well. This setting is practical under the human-in-the-loop configuration suggested by \citet{li-etal-2022-corwa}. However, we realize that this human-in-the-loop setting might be unfriendly to less-experienced researchers who are not sure what keyword to use. Future work may automatically predict the keywords of the gold citation text by performing paragraph- or document-level topic analysis using LLMs.

\paragraph{Choice of ROUGE-labeling approach \& SOTA.} We choose ROUGE-based labeling for CTS because it empirically shows the advantage of using even distantly labeled CTS for citation text generation, compared to using only cited paper abstracts. However, we do not claim that our RAG model using this retriever trained on these distant labels is SOTA. ROUGE-based labeling selects CTS with high lexical overlap with the target citation and may miss CTS that are semantically relevant but less lexically similar. In addition, there are many alternative ways to obtain better performance, such as using a more carefully crafted CTS retriever and training a larger generator with extensive hyper-parameter tuning. Instead, we simply claim that using CTS can improve citation text generation performance over using cited paper abstracts, and human labeling of CTS is not required because distantly labeled CTS is of comparable quality.

\paragraph{Quality Assurance of CTS.} Due to the high cost of manually collecting or evaluating the quality of the retrieved CTS, we are not able to perform quality assurance of the distantly retrieved CTS on a large scale to avoid error-propagation to generated citation texts. Future work can incorporate a fact-checking module to further filter out CTS sentences that do not entail the citation text.

\section*{Ethics Statement}
In this work, we focus on showing that retrieved CTS provides relevant support information to enable faithful citation text generation. Unsurprisingly, the trained CTS-based citation text generation models turn out to be more extractive (Section \ref{sec:results}), directly copying parts of the input CTS to the output citation texts. This copying behavior has the potential to result in plagiarism if the output of our CTS-based citation text generator is directly without further modification. As we mention in Section \ref{sec:results}, it may be necessary to apply a post-processing step, such as a rewriting module, to avoid plagiarism, which is out of the scope of this work and left for future investigation. We also point out that the existing, abstract-based approach to citation text generation extracts longer subsequences from its input, compared to our CTS-based approach, which extracts shorter subsequences and rearranges their order, so the plagiarism problem is not unique to CTS.

\bibliography{anthology,custom}
\bibliographystyle{acl_natbib}

\appendix

\section{More Implementation Details} \label{sec:more_details}
\subsection{Opting Out of FAISS}
The original RAG implementation 
was designed for Wikipedia-based dialogue response generation, where every Wikipedia article is a candidate for retrieval. Therefore, FAISS \cite{johnson2019billion} was used to speed up the search process. However, our search space is limited to cited paper sentences, and we need to retrieve CTS separately for each citation and its corresponding cited paper. Unfortunately, FAISS does not support partitioned search, so we replace FAISS with our simpler CTS retrieval strategies, as described in Section \ref{sec:cts_retreival}.

\subsection{Freezing Query Encoder Weights}
The RAG model retrieves candidate documents on the fly while jointly fine-tuning the pretrained DPR query encoder 
and the generator. 
In our preliminary experiments on retrieving CTS at the paragraph level, we did not find any significant improvement from fine-tuning the query encoder over simply freezing its weights. Further, due to the much larger candidate search space of the sentence-level CTS setting, the on-the-fly retrieval needed to update the query encoder becomes intractable. Therefore, in our citation text generation experiments, we retrieve and cache CTS sentences for all target citation/cited paper pairs in advance and do not fine-tune the query encoder.

\subsection{Citation Text Generation Hyper-parameter Settings}
Table \ref{tab:hyper_parameters} shows the tuned hyper-parameter settings for our CTS-based citation text generation experiments.

\begin{table}[t]
\begin{center}
\small
    \begin{tabular}{  l  l  l }
    \hline
    Hyper-parameter & Search Range & Used \\ \hline
    Learning rate & 1e-5 $\sim$ 5e-4 & 5e-5 \\
    RAG-FiD batch size & 1 $\sim$ 3 & 2 \\
    LED batch size &  - & 1 \\
    Epochs & 2 $\sim$ 10 & 5 \\
    RAG-FiD max sequence length & - & 350 \\
    LED max sequence length & - & 2048 \\
    Top $k$ ROUGE-retrieval & 1 $\sim$ 20 & 10 \\
    \hline
    \end{tabular}
    \vspace{-0.5em}
    \caption{Hyper-parameter settings for CTS-based citation text generation.} \label{tab:hyper_parameters}
    \vspace{-1.5em}
\end{center}
\end{table}

\section{More Related Work}
\subsection{Retrieval and Generation}
Our proposed framework of retrieving useful information (CTS) to guide generation is partly inspired by work in knowledge-augmented, open-domain dialogue generation, where the combination of Dense Passage Retrieval \citep[DPR;][]{karpukhin-etal-2020-dense}, Retrieval-Augmented Generation \citep[RAG;][]{lewis2020retrieval} and Fusion-in-Decoder \citep[FiD;][]{izacard-grave-2021-leveraging} has been shown to yield strong performance \cite{shuster-etal-2021-retrieval-augmentation}.

DPR \cite{karpukhin-etal-2020-dense} separately applies two BERT-like models \cite{devlin-etal-2019-bert} as query and document encoders. 
The DPR model is trained with contrastive loss and uses maximum inner product search (MIPS) to retrieve the documents with the highest cosine similarity between the $[CLS]$ token representations of the query and the knowledge base documents.

Once informative documents have been retrieved, the RAG \cite{lewis2020retrieval} paradigm can be applied using any sequence-to-sequence model. The conditional generation input is paired and encoded with each retrieved document in parallel, and the generation output is jointly decoded using all input/retrieved document pairs. RAG-sequence decodes the entire output sequence based on a fixed distribution over the retrieved documents, while RAG-token decodes each token using a different distribution over the retrieved documents. 
Finally, \citet{shuster-etal-2021-retrieval-augmentation} proposed FiD \cite{izacard-grave-2021-leveraging} as the decoder in the RAG framework by leverageing the attention mechanism to jointly decode from the retrieved document encodings. 

\section{QuestEval and CTS-based Citation Text Generation}
\label{sec:appendix-c}

\begin{table}[t]
\begin{center}
\setlength{\tabcolsep}{3pt} 
\renewcommand{\arraystretch}{1.0} 
\small
    \begin{tabular}{  l | l l l | l l l }
    \hline
     & \multicolumn{3}{l}{\textit{RAG-FiD}} & \multicolumn{3}{l}{\textit{LED}} \\
    \textbf{Setting} & \textbf{R-L} & \textbf{QEval} & \textbf{ANLI} & \textbf{R-L} & \textbf{QEval} & \textbf{ANLI}\\ \hline
    Abstract & 0.556 & \textbf{0.314} & 0.745  & 0.591 & \textbf{0.345} & 0.744  \\ \hline
    Oracle & \textbf{0.726} & 0.288 & 0.940 & \textbf{0.682}  & 0.300 & 0.950 \\ 
    Keyword  & 0.630 & 0.297 & 0.941 & 0.604 & 0.304 & 0.946  \\ 
    Context & 0.641 & 0.310 & \textbf{0.959} & 0.626 & 0.315 & \textbf{0.954}  \\
    \hline
    \end{tabular}
    \vspace{-0.5em}
    \caption{Similarity and faithfulness of the generated citation texts with respect to their input setting, measured by ROUGE-recall, QuestEval, and ANLI.} \label{tab:abstractive_vs_extractive_faithfullness_full}
    \vspace{-1.5em}
\end{center}
\end{table}

QuestEval \cite{scialom-etal-2021-questeval} is a question-answering-based faithfulness metric composed of a precision score, where questions are generated from the summary (citation text) and answered using the document (generation input), and a recall score, where questions are generated from the input and answered using the citation text. Table \ref{tab:abstractive_vs_extractive_faithfullness_full} shows the full faithfulness evaluation results for our citation text generation task, including QuestEval. 

We find that QuestEval encounters some difficulty evaluating CTS-based citation texts because they are very short (Figure \ref{fig:extractiveness} shows that CTS-based citation texts have a much higher compression ratio than abstract-only citation texts); as a result, the QuestEval recall score is extremely low because many questions can be generated from the detailed information in the CTS, but the citation text is too short to contain all of the corresponding answers. 

Table \ref{tab:questeval_example} shows an example of an \textit{Oracle} CTS citation text that receives a low QuestEval score: \citet{wang-etal-2019-best} citing \citet{swayamdipta-etal-2018-syntactic}. We can see that the retrieved CTS sentences include multiple entities and other detailed information that trigger QuestEval to generate many questions that the generated citation text cannot answer. However, a human reader can tell that the \textit{Oracle} citation text is indeed factual; its claim about parse tree validity is supported by \citet{swayamdipta-etal-2018-syntactic}'s Introduction section: \textit{``Since the scaffold task is not an end in itself, we relax the syntactic parsing problem to a collection of independent span-level predictions, with no constraint that they form a valid parse tree."}

We can also see that the \textit{Abstract}-only citation text is longer and contains more entities than the \textit{Oracle} citation text, while the \textit{Abstract} input is much shorter and contains fewer entities than the \textit{Oracle} input, so it does not suffer from the same low QuestEval recall score. We find many similar examples in our test set of longer \textit{Abstract} citations being scored higher than shorter \textit{Oracle} citations by QuestEval.

\begin{table*}[t]
\begin{center}
\setlength{\tabcolsep}{4pt} 
\renewcommand{\arraystretch}{1.0} 
\small
    \begin{tabular}
    {p{0.1\linewidth}p{0.8\linewidth}p{0.05\linewidth}}
    \hline
    \multicolumn{3}{l}{\textbf{Citation context}} \\
    \multicolumn{3}{l}{This information can be incorporated into an SRL system in several different ways.}\\
    \hline
    \multicolumn{3}{l}{\textbf{Citation text}} \\ 
    \emph{Abstract} & \multicolumn{2}{p{0.8\linewidth}}{Swayamdipta et al. (2018) introduced the Syntactic scaffold, an approach to incorporating syntactic information into semantic tasks. The scaffold is designed to avoid the expensive syntactic processing at runtime, only making use of a treebank during training, through a multitask objective, and achieves competitive performance on PropBank semantics, frame semantics, and coreference resolution.}\\
    \emph{Oracle} & \multicolumn{2}{p{0.8\linewidth}}{Swayamdipta et al. (2018) use syntactic information to predict spans, but it is not a guarantee that a span is a valid parse tree.}\\
    \hline
    \multicolumn{3}{l}{\textbf{Cited text spans}} \\
    
    \emph{Abstract} & \multicolumn{2}{p{0.8\linewidth}}{We introduce the syntactic scaffold, an approach to incorporating syntactic information into semantic tasks. Syntactic scaffolds avoid expensive syntactic processing at runtime, only making use of a treebank during training, through a multitask objective. We improve over strong baselines on PropBank semantics, frame semantics, and coreference resolution, achieving competitive performance on all three tasks.}\\
    \emph{Oracle} & \multicolumn{2}{p{0.8\linewidth}}{Span Features: • width of the span in tokens (Das et al., 2014) • distance (in tokens) of the span from the target • position of the span with respect to the target (before, after, overlap) Each of these features is encoded as a one-hot embedding and then linearly transformed to yield a feature vector, a i: j . \newline
    \textit{Introduction: Since the scaffold task is not an end in itself, we relax the syntactic parsing problem to a collection of independent span-level predictions, with no constraint that they form a valid parse tree.} \newline
    Latent variables.: The motivation behind joint learning of syntactic and semantic representations is that any one task is helpful in predicting the other (Lluís and Màrquez, 2008; Lluís et al., 2013; Henderson et al., 2013; Swayamdipta et al., 2016) . \newline
    Semi-Markov CRF: In order to model the non-overlapping arguments of a given target, we use a semi-Markov conditional random field (semi-CRF; Sarawagi et al., 2004) . \newline
    Coreference Resolution: Labels s 1 to s m indicate a coreference link between s and one of the m spans that precede it, and null indicates that s does not link to anything, either because it is not a mention or it is in a singleton cluster. \newline
    Related Work: Using syntax as T 2 in a pipeline is perhaps the most common approach for semantic structure prediction (Toutanova et al., 2008; Yang and Mitchell, 2017; Wiseman et al., 2016) . \newline
    Coreference Resolution: Since there is a complete overlap of input sentences between D sc and D pr as the coreference annotations are also from OntoNotes (Pradhan et al., 2012), we reuse the v for the scaffold task. \newline
    Results: Tan et al. (2018) employ a similar approach but use feed-forward networks with selfattention. \newline
    Results: Prior competitive coreference resolution systems (Wiseman et al., 2016; Clark and Manning, 2016b,a) all incorporate synctactic information in a pipeline, using features and rules for mention proposals from predicted syntax. \newline
    Results: We follow the official evaluation from the SemEval shared task for frame-semantic parsing (Baker et al., 2007) .}\\
    \hline
    \end{tabular}
    \vspace{-0.5em}
    \caption{Citation text generation example, \citet{wang-etal-2019-best} citing \citet{swayamdipta-etal-2018-syntactic}, showing the difference in compression ratio between the \textit{Abstract} and \textit{Oracle CTS} settings that results in unexpected QuestEval recall scores. The \textit{Oracle} generation is supported by the \textit{italicized} CTS sentence but still receives a low QuestEval score.} \label{tab:questeval_example}
    \vspace{-1.5em}
\end{center}
\end{table*}

\label{sec:appendix-questeval}

\end{document}